\documentclass[letterpaper, 10 pt, conference]{ieeeconf}
\makeatletter
\let\MYcaption\@makecaption
\makeatother

\usepackage[font=footnotesize]{subcaption}

\makeatletter
\let\@makecaption\MYcaption
\makeatother

\usepackage{graphics} 
\usepackage{epsfig} 
\usepackage{amsmath} 
\usepackage[T1]{fontenc}
\usepackage{bm}
\usepackage{cite}
\usepackage{here}
\usepackage{booktabs}
\usepackage{threeparttable}
\usepackage{multirow}
\usepackage{color} 
\usepackage{cases}
\usepackage{siunitx}

\usepackage{algorithm}
\usepackage{algpseudocode}
\usepackage{makecell}
\usepackage{outlines}
\usepackage{amssymb}

\usepackage{float} 
\newtheorem{assumption}{Assumption}

\usepackage{url}
\usepackage{tabularray}
\usepackage[normalem]{ulem}

\newcommand{\slfrac}[2]{\left.#1\middle/#2\right.}

\IEEEoverridecommandlockouts                              
\title{\LARGE \bf 
A Unified Interaction Control Framework for Safe Robotic \\ Ultrasound Scanning with Human-Intention-Aware Compliance
}
\author{Xiangjie Yan, Shaqi Luo, Yongpeng Jiang, Mingrui Yu, Chen Chen, \\
Senqiang Zhu, Gao Huang, Shiji Song and Xiang Li
\thanks{X. Yan, Y. Jiang, M. Yu, C. Chen, G. Huang, S. Song and X. Li are with Department of Automation, Beijing National Research Center for Information Science and Technology, Tsinghua University. S. Luo is with Beijing Academy of Artificial Intelligence. S. Zhu is with Midea Corporate Research Center and State Key Laboratory of High-end Heavy-load Robots, Midea Group. This work was supported in part by the Science and Technology Innovation 2030-Key Project under Grant 2021ZD0201404, in part by the National Natural Science Foundation of China under Grant U21A20517 and 52075290, in part by the Institute for Guo Qiang, Tsinghua University, in part by the State Key Laboratory of High-end Heavy-load Robots under Grant HHR2024010426, and in part by Beijing Natural Science Foundation under Grant QY23121.  Corresponding author: Xiang Li (xiangli@tsinghua.edu.cn)}
}

\begin{document}
\maketitle
\pagestyle{empty}  
\thispagestyle{empty} 

\begin{abstract}
The ultrasound scanning robot operates in environments where frequent human-robot interactions occur. 
Most existing control methods for ultrasound scanning address only one specific interaction situation or implement hard switches between controllers for different situations, which compromises both safety and efficiency.
In this paper, we propose a unified interaction control framework for ultrasound scanning robots capable of handling all common interactions, distinguishing both \textit{human-intended} and \textit{unintended} types, and adapting with appropriate compliance.
Specifically, the robot suspends or modulates its ongoing main task if the interaction is \textit{intended}, e.g., when the doctor grasps the robot to lead the end effector actively.
Furthermore, it can identify \textit{unintended} interactions and avoid potential collision in the null space beforehand. Even if that collision has happened, it can become compliant with the collision in the null space and try to reduce its impact on the main task (where the scan is ongoing) kinematically and dynamically.
The multiple situations are integrated into a unified controller with a smooth transition to deal with the interactions by exhibiting human-intention-aware compliance. 
Experimental results validate the framework's ability to cope with all common interactions including intended intervention and unintended collision in a collaborative carotid artery ultrasound scanning task.

\end{abstract}

\section{Introduction}
Ultrasound scanning is a common noninvasive health screening method in great demand. 
During an ultrasound scanning task, a doctor holds a scanning probe and moves it along a patient's body, maintaining close and stable contact to enable clear ultrasound imaging. Ultrasound scanning is labor-intensive, as doctors are typically required to perform several hours of imaging per day \cite{ardms}. Therefore,
deploying a robot to carry out these scanning tasks autonomously helps alleviate this problem \cite{hennersperger2016towards,kojcev2016dual,8002599,jiang2024cardiac}. To achieve this, various ultrasound scanning robots have been developed \cite{Li2021,welleweerd2021out,dyck2022impedance}, and many works have been devoted to the navigation systems \cite{li2021image,bi2022vesnet}, which output desired probe trajectory to obtain the ideal ultrasound image with the current ultrasound image and robot information as input. However, most works only consider the navigation while ignoring the physical interaction between the scanning robot and the human, which is an essential problem for safe ultrasound scanning.

\begin{figure}[tb]
\centering
\includegraphics[width=\linewidth]{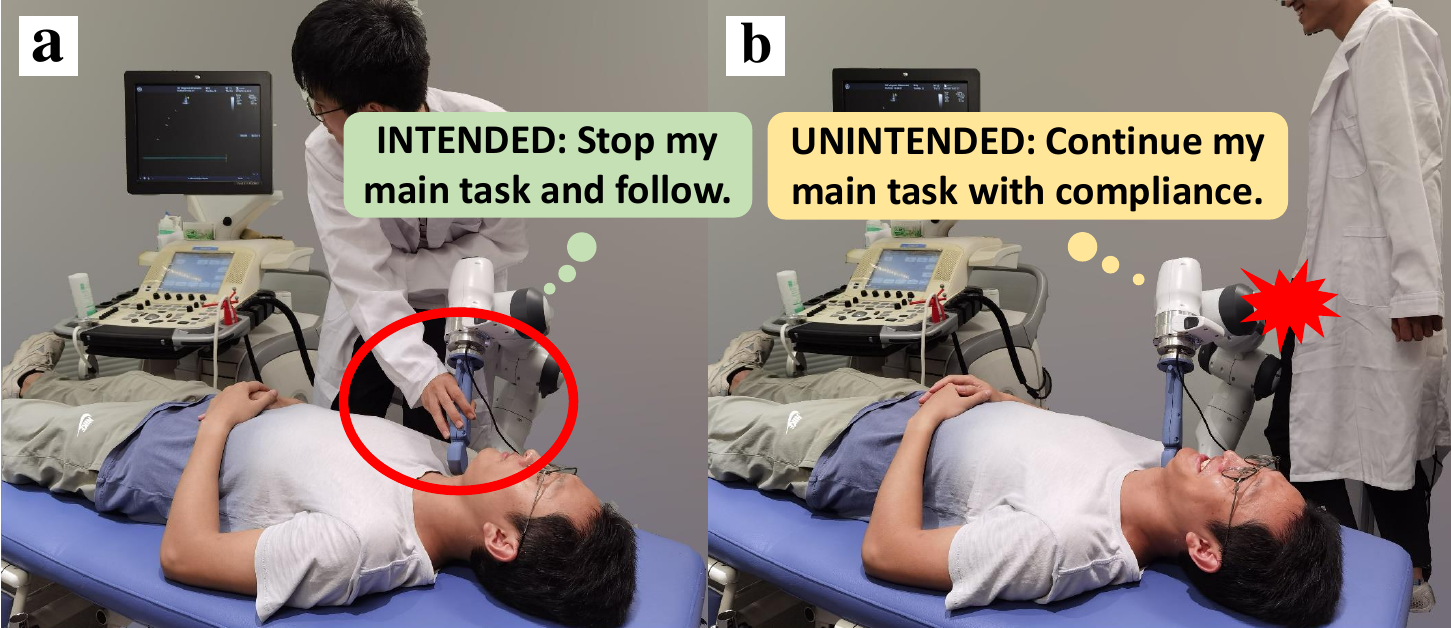}
\caption{An illustration of the human-intention-aware compliance during a carotid artery examination by an ultrasound scanning robot. (a) The doctor grasped the probe to apply coupling gel on the probe, and the robot stopped its current task and followed. (b) The doctor collided with the robot by his leg by accident, and the robot continued its main task with compliance.}\label{fig_scene}
\vspace{-6mm}
\end{figure}

In addition to the normal contact between the scanning probe and the patient's skin, there are many types of physical human-robot interactions commonly happening during the robot-assisted scanning, such as patients moving, doctors intervening, and doctors colliding.
These interactions can be categorized into two types. i) {\em Human-Intended Interaction}: e.g., the doctor holds the robot end effector and actively adjust its motion (Fig.~\ref{fig_scene}a); ii) {\em Human-Unintended Interaction}: e.g., the doctor may carelessly collide with the robot body, resulting in unexpected impacts (Fig.~\ref{fig_scene}b). Different robot reactions are required for different human-robot interactions during the ultrasound scanning process.

Some existing works have applied typical interaction control methods to robotic ultrasound scanning, such as using optimization-based control \cite{duan2022ultrasound}, hybrid force/position control \cite{goel2022autonomous}, or impedance/admittance control \cite{fu2024optimization, dyck2022impedance, finocchi2017co}. However, most works only consider the interaction between the robot and patient, neglecting the intervention from the doctor. 
Moreover, they usually separately dealt with single interaction situation, without achieving a unified solution for all potential interactions. 
Hard switching \cite{7759101} between different controllers for different interaction situations will affect the smoothness and safety of the scanning process.

To address the problems, this paper proposes a unified interaction control framework for safe robotic ultrasound scanning with human-intention-aware compliance, where the robot's working mode varies according to the changing human intention. It adapts to the human-intended interactions and rejects the human-unintended ones.
Specifically, this framework unified achieves the following functions:
\begin{itemize}
    \item The robot will comply with the human-intended interactions. That is, when the doctor or patient grasps the probe and actively moves it, the robot will compliantly follow the human's guidance. 
    
    \item The robot will try to actively avoid human-unintended collisions between the human and robot body without affecting the main scanning task executed by the end-effector, which improves safety.
    
    \item The robot will become compliant to the external human-unintended impact when the collision or impact is unavoidable and actually happens. The achieved compliance will reduce the unsafe impact on the human without affecting the main scanning task.
\end{itemize}

The above functions are modularized as different working modes, which are integrated into a unified control framework with smooth transitions.
Such a formulation is able to deal with all common physical interactions during the human-robot collaborative ultrasound scanning in a unified and smooth manner, making the human's knowledge and the robot's ability complement each other efficiently and safely.
The performance of the proposed framework is validated in real-world carotid artery examinations.
The key contribution of this work is summarized as follows:
\begin{itemize}
    \item We proposed a unified control framework considering all common human-robot interactions in the human-robot collaborative ultrasound scanning task.
    \item We formulated different types of interactions into several modes and proposed a smooth human-intention-aware mode transition approach with weighting factors.
    \item We developed an integrated robotic system that achieved reliable human-robot collaborative ultrasound scanning for human carotid arteries in the real world.
\end{itemize}

\section{Related works}
\subsection{Ultrasound Scanning Robots}
As mentioned in the Introduction, lots of works for ultrasound scanning robots focused on navigation rather than physical interaction with humans, including intended and unintended ones, which is important for safety in a crowded clinical environment and efficiency when the human wants to intervene. The ethical and legal regulations for clinical translation and further commercialization of ultrasound scanning robots underline the importance of such interactions, that is, there must be at least one human supervisor for autonomous ultrasound scanning systems (see IEC/TR 60601-4-1 \cite{IEC} formulated by \textit{the International Organization for Standardization (ISO)} and \textit{the International Electrotechnical Commission (IEC)}). This paper aims to further reinforce the doctor's role of supervisor, by allowing him or her to safely intervene anytime during the scanning task, with the development of a new interaction control scheme.

Moreover, most controllers for ultrasound scanning robots are designed for specific situations. A shared control method was proposed in \cite{selvaggio2021autonomy} to allow both the doctor and the robot to contribute to the probe movement. That is, the doctor first moves the probe to a rough position, and the robot, regulated by an image-based visual servoing method, works to track an existing feature in ultrasonic image autonomously \cite{nakadate2010implementation,hennersperger2016towards,kojcev2016dual,8002599}. In \cite{7759101}, authors proposed a hybrid force/position regulation controller for contacting and switched to a position controller for out-of-plane rotation in an aortic diameter measurement task. These works didn't consider one control framework for all common interaction situations. In this paper, a unified control framework and smooth transition for different working modes are achieved.

\subsection{Interaction Control}
Impedance control is a critical interaction control method, particularly in redundant robots. 
Hierarchical impedance control, including Operational Space Formulation \cite{khatib1987unified, nakanishi2008operational, dietrich2018hierarchical} and hierarchical compliance control \cite{ott2008resolving,ott2015prioritized}, is widely used to manage robot-environment interactions.
Null-space impedance control schemes, such as those in \cite{sadeghian2011multi}, enable multiple prioritized tasks at the acceleration level. Extended task space concepts \cite{oh1998extended, sadeghian2013task} address null-space interactions without requiring joint-torque measurements, aided by generalized force observers. 
In \cite{kronander2015passive}, a controller was proposed to ensure passivity in trajectory-tracking tasks. In \cite{schindlbeck2015unified}, contact-loss stabilization was guaranteed by combining force tracking and impedance control. These approaches primarily focused on flat or regular surfaces. In contrast, \cite{dyck2022impedance} was able to deal with complex arbitrary surfaces in task space. Moreover, energy-tank-augmented methods\cite{ferraguti2019variablea} fixed the deterioration of passivity caused by variable impedance operation and null-space projection operation\cite{dietrich2016passive,michel2022safety}. The theoretical guarantee of stability and passivity helps to build a safe unified control framework in this paper.

\section{Preliminaries}
\subsection{Task Definitions and Robot Dynamics}
The main and secondary task coordinates and their velocities of a redundant robot can be expressed as 
\begin{align}
\bm x_1=\bm f_1(\bm q),\quad \bm x_2=\bm f_2(\bm q),\\ 
\dot{\bm x}_1=\bm J_1(\bm q)\dot{\bm q},\quad \dot{\bm x}_2=\bm J_2(\bm q)\dot{\bm q},\label{velEqn}
\end{align}
where $\bm q\hspace{-0.05cm}\in\hspace{-0.05cm}\Re^n$ is the vector of joint angles, $n$ is the number of DOFs for a redundant robot, \(\bm x_i \in \Re^{m_i}, i=1,2\) means the main and secondary task, respectively, \(\bm f_i: \Re^n \xrightarrow{} \Re^{m_i}\) are differentiable forward kinematic functions, $\bm J_i(\bm q)\hspace{-0.05cm}\in\hspace{-0.05cm}\Re^{m_i\times n}$ are the Jacobian matrices. The overall task dimension \(m=\Sigma^2_{i=1} m_i=n\).
The augmented task-space velocity \(\dot{\bm x}\in \Re^n\) is
\begin{equation}
    \dot{\bm x}:=\left[\begin{array}{c}
\dot{\bm x}_1 \\
\dot{\bm x}_2
\end{array}\right]=\bm J(\bm q) \dot{\bm q},
\end{equation}
where the augmented Jacobian matrix is
\begin{eqnarray}
    \bm J(\bm q):=\left[\begin{array}{c}
\bm J_1(\bm q) \\
\bm J_2(\bm q)
\end{array}\right]\in \Re^{n\times n}.
\end{eqnarray}
A common and practical assumption \cite{ott2008resolving, ott2015prioritized} is made to simplify the analysis:
\begin{assumption}
The main task and secondary task are independent and do not have singularities, i.e., \(rank(\bm J_1)+rank(\bm J_2)=m_1+m_2=n\).\label{ass1}
\end{assumption}

The ultrasonic scanning robot involves rich contact with doctors and patients, its dynamic model is given as
\begin{eqnarray}
&\bm M(\bm q)\ddot{\bm q}+\bm C(\dot{\bm q}, \bm q)\dot{\bm q}+\bm g(\bm q)=\bm \tau+\bm \tau_{e},\label{eq_dyn}
\end{eqnarray}
where \(\bm M(\bm q)\hspace{-0.05cm}\in\hspace{-0.05cm}\Re^{n\times n},\bm C(\dot{\bm q}, \bm q)\dot{\bm q}\hspace{-0.05cm}\in\hspace{-0.05cm}\Re^n,\bm g(\bm q)\hspace{-0.05cm}\in\hspace{-0.05cm}\Re^n\) denote the mass matrix,  Coriolis and centrifugal term, and gravity vector respectively, and \(\bm \tau, \bm \tau_e\hspace{-0.05cm}\in\hspace{-0.05cm}\Re^n\) represent the control torque and the external torque respectively. Note that both the contact between the scanning probe and the patient's body and the collision with the robot contributes to the external torque.

\subsection{Hierarchically Decoupled Task and Dynamics}\label{preliminarier_decoupled_dynamics}
Since the original task coordinates are coupled between different levels, this causes difficulties in hierarchical control and stability analysis. Hierarchically decoupled task-space velocities \(\bm v_i\in \Re^{m_i}, i\in \{1,2\}\) are defined as\footnote{The dependencies are omitted in the rest of the paper for simplicity.}
\begin{align}
  \bm v_i:=\bar{\bm{J}}_i \dot{\bm q},
\end{align}
where \(\bar{\bm{J}}_i\) is the decoupled Jacobian:
\begin{small}
\begin{align}
\bar{\bm{J}}_1&=\bm{J}_1, \\
\bar{\bm{J}}_1^{M+} & =\bm M^{-1} \bar{\bm{J}}_1^T(\bar{\bm{J}}_1 \bm{M}^{-1} \bar{\bm{J}}_1^T)^{-1}, \label{eq_dyn_con_inverse}\\
\bar{\bm{J}}_2&=(\bm Z_2 \bm M \bm Z_2^T)^{-1}\bm Z_2 \bm M,\label{eq_j2ba}
\end{align}
\end{small}
where (\ref{eq_dyn_con_inverse}) is the well-known dynamically consistent inverse \cite{khatib1987unified, dietrich2015overview}. 
In (\ref{eq_j2ba}), we need to know the full row-rank null-space base matrix \(\bm Z_2\in \Re^{m_2\times n}\), where \(\bm Z_2\) spans the null space of \(\bm J_1\), i.e., \(\bm J_1 \bm Z_2^T=\bm 0\). Such a matrix \(\bm Z_2\) can be obtained by many ways like singular value decomposition or recursive methods.
Here we use singular value decomposition: \(\bm J_1 = \bm U \bm \Sigma \bm V^T, \bm V=[\bm X_1^T, \bm Y_1^T]\). We can define \(\bm Z_1 = \bar{\bm{J}}_1^{M+T}=\bm{J}_1^{M+T}, \bm Z_2 = \bm Y_1\). Such a definition of \(\bm Z_i\) also fulfills
\begin{small}
\begin{align}
\bm{Z}_1 \bar{\bm{J}}_1^T = \bm I, \quad &\bm{Z}_2 \bar{\bm{J}}_2^T = \bm I,\label{eq_10}\\
\bm{Z}_1 \bm{M} \bm{Z}_2^T & =\mathbf{0}\label{eq_11}.
\end{align}
\end{small}
Now, the augmented task velocity in decoupled coordinates can be defined as
\begin{equation}
    \bm v:=\left[\begin{array}{c}
\bm v_1 \\
\bm v_2
\end{array}\right]
=\bar{\bm J}\dot{\bm q}, \bar{\bm J}:=\left[\begin{array}{c}
\bar{\bm{J}}_1 \\
\bar{\bm{J}}_2
\end{array}\right].\label{eq_12}
\end{equation}
According to assumption \ref{ass1}, \(\bar{\bm J}\) is also non-singular. Using the properties (\ref{eq_10})(\ref{eq_11}), a useful property can be obtained that \(\bar{\bm J}^{-1} = [\bm Z_1^T, \bm Z_2^T]\).
The decoupled robot dynamics \cite{ott2015prioritized} is 
\begin{equation}
    \bm{\Lambda} \dot{\bm{v}}+\bm{\mu} \bm{v}=\bar{\bm{J}}^{-T}\left(\bm{\tau}+\bm{\tau}_e-\bm{g}\right),
\end{equation}
where
\begin{small}
\begin{align}
& \bm{\Lambda}=\bar{\bm{J}}^{-T} \bm{M} \bar{\bm{J}}^{-1}=\text{diag}(\bm\Lambda_{1}, \bm\Lambda_{2}), \\
& \bm{\mu}=\left(\bar{\bm{J}}^{-T} \bm{C}-\bm{\Lambda} \dot{\bar{\bm{J}}}\right) \bar{\bm{J}}^{-1}=
\left[\begin{array}{cc}
\bm{\mu}_{11} & \bm{\mu}_{12} \\
\bm{\mu}_{21} & \bm{\mu}_{22}
\end{array}\right].
\end{align}
\end{small}

\section{Methodology}
\begin{figure}[!tb]
\centering
\includegraphics[width=8cm]{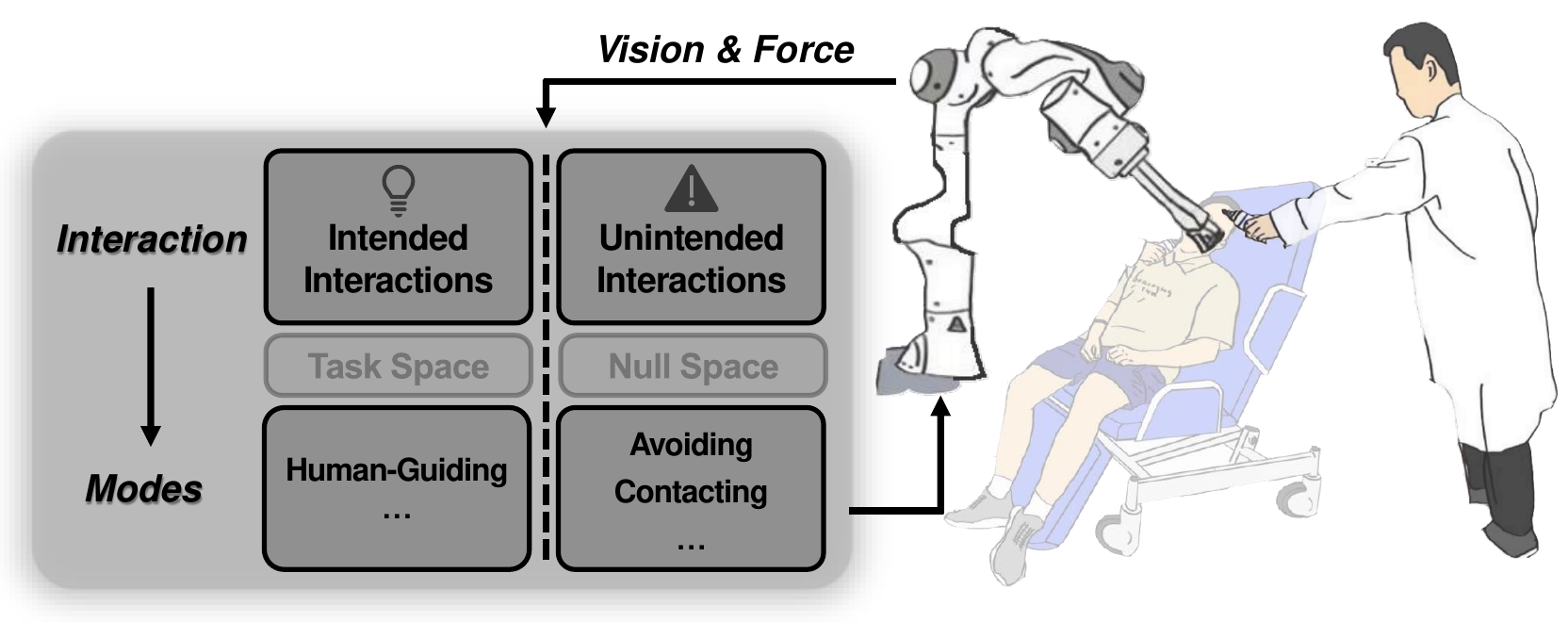}
\vspace{-0.3cm}
\caption{The human-robot interactions considered in this work. The interactions are divided into intended- and unintended- situations and are further classified as several modes based on vision and force information.}\label{fig_interaction_situations}
\vspace{-0.3cm}
\end{figure}

We propose a unified interaction control method for ultrasound scanning robots to handle complex and dynamic interactions during scanning tasks.
If the interaction is human-intended, adjust the main task to accommodate the interaction; if the interaction is unintended, only the null-space task is adapted to it without affecting the main task, as shown in Fig. \ref{fig_interaction_situations}.

\begin{figure}[!tb]
\centering
\includegraphics[width=8cm]{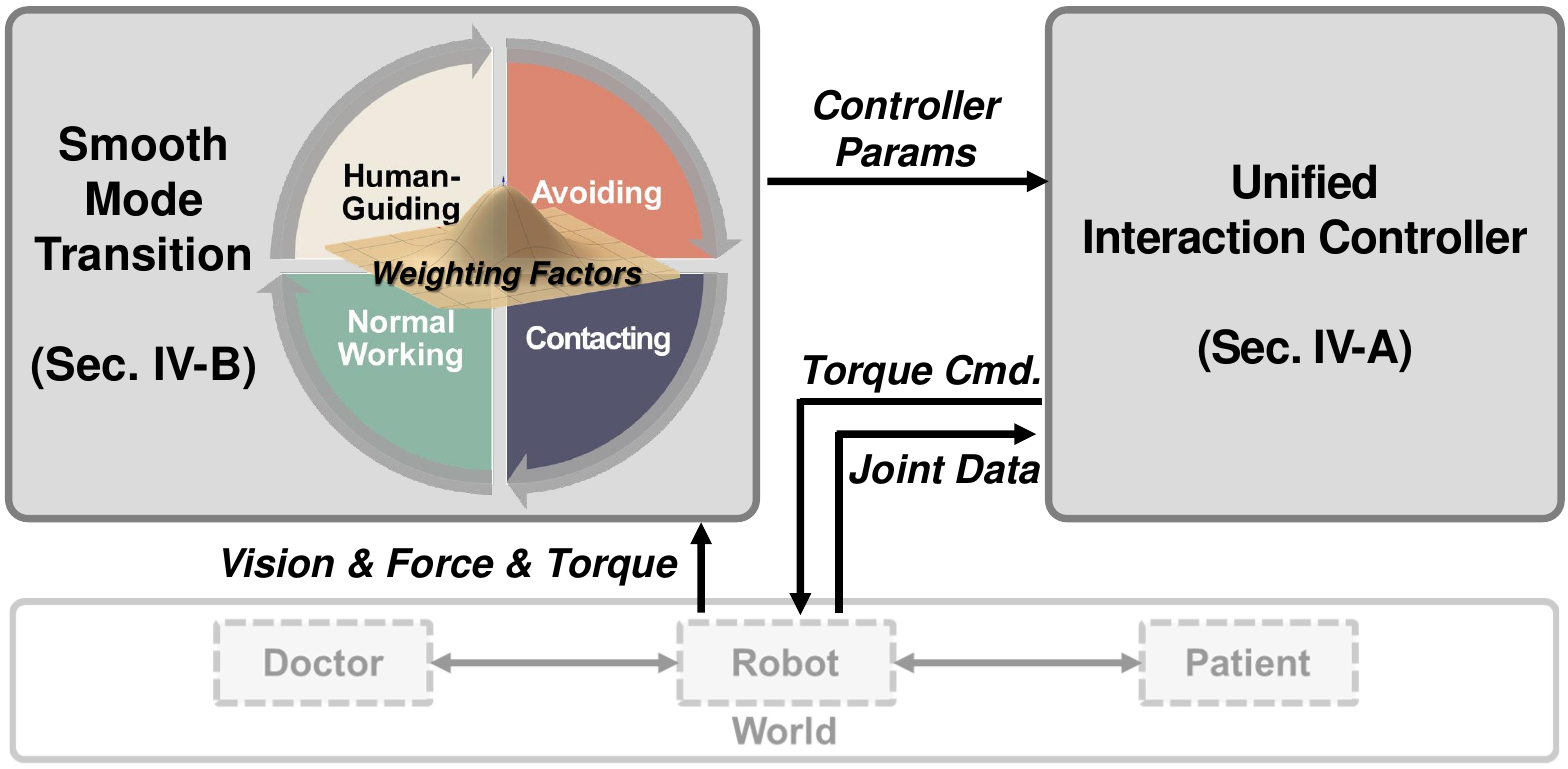}
\vspace{-0.3cm}
\caption{The structure of our proposed method, where multiple modes are integrated into a unified control input with smooth transitions to deal with the interactions by exhibiting human-intention-aware compliance.}\label{fig_structure}
\vspace{-0.6cm}
\end{figure}

\begin{table*}[!htb]
\centering
\caption{Potential physical interactions during the scanning and the designed corresponding robot working mode.}
\label{table_mode}
\vspace{-0.2cm}
\begin{tblr}{
  row{1} = {c},
  cell{1}{1} = {c=2}{},
  cell{2}{1} = {r=3}{},
  cell{3}{2} = {r=2}{},
  cell{5}{1} = {c=2,r=3}{c},
  hline{1-2,5,8} = {-}{},
  hline{3} = {2-5}{},
  hline{4,6-7} = {3-5}{},
}
FUNCTION                 &            & MODE                                  & ROBOT BEHAVIOR                                                            & ACTIVATED WHEN                                                                                 \\
{Handling\\Interactions} & Intended   & \em Human-Guiding Mode & Follow
the human’s
movement.                                  & Someone
grasps
the
probe.                                                             \\
                         & Unintended & \em Avoiding Mode      & Avoiding without affecting scanning task.                     & The doctor is too
close to
the robot body.                                          \\
                         &            & \em Contacting Mode    & {Compliant with the contact~without \\affecting the scanning task.} & {The
doctor
unintentionally \\contacts the
robot body.}                             \\
{Normal\\Working }       &            & \em Recovery Mode      & Recontact the patient’s
body.                                 & The
patient loses
contact with
the probe.                                             \\
                         &            & \em Scanning Mode      & {Keep moving along the scanning \\trajectory.}                & {The
probe
contacts with the
patient's
neck, \\and the scanning trajectory is given.} \\
                         &            & \em Waiting Mode       & Keep static.                                                   & The scanning trajectory is unknown.                                                   
\end{tblr}
\vspace{-0.2cm}
\end{table*}

\begin{table*}[!tb]
\centering
\caption{Mode Design}
\label{table_mode}
\vspace{-0.2cm}
\begin{tblr}{
  cells = {c},
  cell{2}{3} = {r=6}{},
  hline{1,8} = {-}{0.08em},
  hline{2} = {-}{0.05em},
  hline{3-7} = {1-2}{},
}
Mode                                      & Desired values in the unified controller (\ref{eq_tau1_regu})(\ref{eq_tau2_regu})                                                                                                                                                   & Impedance params \\
{\em Recovery Mode}      & {$\bm x_{1d},\dot{\bm x}_{1d},\ddot{\bm x}_{1d}$ are given by a mini-jerk trajectory generator, use controller (\ref{eq_tau1_tracking})(\ref{eq_tau2_regu});\\$\bm x_{2d}$ is updated to actual robot value every 100 control cycles;} &{$\bm K_1 = (1-a_h)$\\$(1-a_p)(1-a_f)\bm K_{1g},$\\ $\bm K_2 = (1-a_h)(1-a_n)\bm K_{2g},$\\Critical damping.}           \\
{\em Human-Guiding Mode} & $\bm x_{1d}, \bm x_{2d}$ are set as current pose, i.e., $\bm x_{1d}(t) = \bm x_1(t), \bm x_{2d}(t) = \bm x_2(t)$);                                                                                 &           \\
{\em Scanning Mode}      & $\bm x_{1d}, \bm x_{2d}$ are given by the scanning trajectory generator in our previous work \cite{yan2023multi};                                                                                                                &           \\
{\em Avoiding Mode}      & \(\bm x_{1d}\) is the same with {\em Scanning Mode}, \(\bm x_{2d}{\mathrel{+}=}a_b \Delta\);                                                                                                                 &           \\
{\em Contacting Mode}    & \(\bm x_{1d},\bm x_{2d}\) are the same with {\em Scanning Mode};                                                                                                                 &           \\
{\em Waiting Mode}       & $\bm x_{1d}, \bm x_{2d}$ is fixed to the pose at the moment entering this mode;                                                                                                                     &           
\end{tblr}\vspace{-0.2cm}
\end{table*}
For this purpose, we first conclude several most common types of interaction situations during a scanning task, which are shown in the left part of Fig. \ref{fig_structure} and more detailed in Tab. \ref{table_mode}. Three specific working modes are designed for handling interactions:
\begin{itemize}
    \item {\em Human-Guiding Mode}: The robot follows the human's movement when the probe is grasped.
    \item {\em null-space Avoiding Mode}\footnote{the term ``{\em null-space}'' will be neglected in the following for simplicity.}: The robot adjusts its redundant joints to avoid the human when they are too close, without affecting the scan.
    \item {\em null-space Contacting Mode}: when the doctor unintentionally contacts the robot body, the robot is compliant with the contact without affecting the scanning task.
\end{itemize}
The other three modes are for normal working, which is the obligated for an ultrasound robot:
\begin{itemize}
    \item {\em Scanning Mode}: The robot follows the scanning trajectory when the probe is in contact.
    \item {\em Waiting Mode}: The robot remains stationary when the trajectory is unknown.
    \item {\em Recovery Mode}: The robot re-establishes contact when the probe loses connection with the patient.
\end{itemize}

Smooth transitions between modes are critical. Existing methods rely on hard switching \cite{huang2018fully,duan2022ultrasound}, jeopardizing the stability and, hence, the safety of the human. In contrast, our method takes three steps to achieve that. First, we modified the hierarchical compliance control, with changing parameters to achieve different modes. Second, we bridge the changing parameters with perception data using smooth weighting factors. Third, we are able to show that this framework remains passive.

\subsection{Unified Interaction Controller Based on Hierarchical Compliance Control}\label{sec_controller}

We consider a redundant robot configuration which provides more flexibility for crowded hospital environment. In this paper, we use a 7-DOF manipulator and consider a 2-level hierarchy: The first level task is the end effector pose, \(m_1=6\), which is vital in the scanning task. The secondary task is the angle of joint 1, \(m_2=1\). Not only does such a choice meet assumption 1, but also the scalar format of the task 2 coordinate\footnote{Although task 2 coordinate is a scalar in this task, we still use boldface to represent this scalar as a vector in order to show the generality of the theory.
} makes it easier to manipulate the null-space configuration compared to other common formats like 7 DOF joint impedance.

Here, we summarize the hierarchical compliance control method \cite{ott2008resolving, ott2015prioritized}, which ensures asymptotic stability and desired impedance behavior.

The controller can be expressed as
\begin{align}
    \bm \tau=\bm g+\bm \tau_d+ \bm \tau_1+ \bm \tau_2-\bm \tau_{e},\label{eq_controller}
\end{align}
where \(\bm \tau_d\) is the compensation term:
\begin{equation}
\bm{\tau}_d=\bar{\bm{J}}_1^T \bm{\mu}_{12} \bm{v}_2+\bar{\bm{J}}_2^T \bm{\mu}_{21} \bm{v}_1, 
\end{equation}
one can obtain a property \(\bm \tau^T_d \dot{\bm{q}}\hspace{-0.05cm}=\hspace{-0.05cm}\bm 0\), indicating zero transmitted power with respect to the effort-flow pair \((\bm \tau_d, \dot{\bm q})\) \cite{ott2008resolving}. 
The last two terms in (\ref{eq_controller}) are the control torque for each subtask, which can be written as
\begin{gather}
    \bm \tau_1 = \bar{\bm{J}}_1^T \left(-\bm K_{1}\tilde{\bm{x}}_1-\bm D_1 \dot{\bm{x}}_1\right),\label{eq_tau1_regu}\\
    \bm \tau_2 = \bar{\bm{J}}_2^T \bm Z_2 \bm{J}_2^T\left(-\bm K_{2} \tilde{\bm{x}}_2-\bm D_2 \dot{\bm x}_2\right ) , \label{eq_tau2_regu}
\end{gather}
where \(\tilde{\bm{x}}_i=\bm{x}_i-\bm{x}_{id}\) is the error of the \(i\)th task, \(\bm{x}_{id}\) denotes the desired value of the \(i\)th task. The stiffness and damping components are represented by matrices \(\bm K_i\) and \(\bm D_i\), respectively. Note that these impedance parameters are time-varying during the scanning task, and hence able to be in different modes.

In {\em Recovery Mode}, when the probe is far away from the patient's neck, the 
min-jerk trajectory generator is utilized to give the desired trajectory from the current pose to the desired pose. Unlike other modes, a tracking task rather than a regulation task is requested in this mode, so we need to slightly change our controller (\ref{eq_tau1_regu}) into the tracking form:
\begin{equation}
    \bm \tau_1 = \bar{\bm{J}}_1^T (\ddot{\bm{x}}_{1d}-\bm K_{1}\tilde{\bm{x}}_1-\bm D_1 \dot{\tilde{\bm{x}}}_1).\label{eq_tau1_tracking}
\end{equation}

Next, we will use the design of the mode identification and transition to adapt the hierarchical compliance controller to the contact-rich and dynamically interactive ultrasound scanning task with the realization of compliance that is indicated by human intention.

\subsection{Mode Design and Smooth Transition}
The connection between different working modes is managed using five weighing factors in \([0,1]\), each representing an aspect of the task:
\begin{enumerate}
    \item \(a_h\): if the probe is grasped by a human \textbf{h}and.\footnote{\(a_h=1\) for true, \(0\) for false, the same for other factors. The bold letters show the meaning of the subscripts};
    \item \(a_p\): if the probe is near the \textbf{p}atient's neck;
    \item \(a_f\): if the contact between the probe and the patient's neck is maintained, based on \textbf{f}orce data;
    \item \(a_n\): if the doctor contacts the robot's body, causing a \textbf{n}ull-space force;
    \item \(a_b\): if the doctor contacts the robot's \textbf{b}ody, based on position data.
\end{enumerate}

To ensure a smooth transition, we define a basic function:
\begin{subnumcases}{b(s)=}\slfrac{1}{1+s^6}\,, & \(s\geq 0\), \\
    1\,, &\(s<0\),
\end{subnumcases}
The basic function is a continuous transition from $1$ to $0$, where $s$ is a variable. The weighting factors are constructed through this basic function and perception data.
The design of all weighting factors is detailed below.

The first and the last weighting factors $a_h, a_b$ use position information obtained from the perception system:
\begin{gather}
    a_h = b\biggl(\frac{ d_h }{r_h}\biggr),
    a_b = b\biggl(\frac{d_{b}}{r_b}\biggr),
\end{gather}
where \(d_h\) is the distance from the probe to one's hand and \(d_b\) is the closest distance from the robot's body to the doctor's body\footnote{in the real-world experiment, ``the doctor's body'' is specified as the doctor's hands.}, the two scalars \(r_h, r_b\) are scaling parameters.

The second weighting factor also uses position information, modeling the patient’s neck as a cylindrical region:
\begin{equation}
    a_p = b\Biggl(\frac{\sqrt{d_{py}^2+d_{pz}^2}}{r_p}\Biggr)\cdot b \Biggl( \frac{\lvert d_{px} - \frac{x_{\text{top}}+x_{\text{bottom}}}{2} \rvert}{\frac{x_{\text{top}}-x_{\text{bottom}}}{2}} \Biggr),
\end{equation}
where \(\bm d_p = [d_{px}, d_{py}, d_{pz}]^T\) is the position vector of the probe in the neck frame, \(r_p\) is the radius of the region, and \(x_{\text{top}}, x_{\text{bottom}}\) are the x coordinates of the top and bottom surfaces of the region, respectively.

The third weighting factor is:
\begin{equation}
    a_f = 1 - b\biggl(\frac{^{E}f_z}{f_0}\biggr),
\end{equation}
where $f_0$ is a threshold constant, and $^Ef_z$ denotes the force along the $z$-axis in the end effector frame $\{E\}$. Maintaining steady contact during scanning requires $^Ef_z\geq0$. 

The fourth weighting factor \(a_n\) is defined as:
\begin{align}
    a_n = 1 - b\biggl(\frac{|\bm\tau_n|}{\tau_0}\biggr),\quad \bm \tau_n = \bm \tau_e-\bm J_1^T \bm F_{1e},
\end{align}
where \(\tau_0\) is a null-space torque threshold. \(\bm \tau_n\) is the null-space torque where \(\bm F_{1e}\in \Re^{6\times 1}\) is the external wrench detected by the wrist-mounted F/T sensor.

The stiffness and desired task coordinates vary according to these five scalars to achieve the behaviors, as shown in Tab. \ref{table_mode}, 
The unified stiffness is designed as:
\begin{align}
    \bm K_{1}(t)&= (1-a_h)(1-a_f)(1-a_p) \bm K_{1g},\label{eq_update1}\\
    \bm K_{2}(t)&= (1-a_h)(1-a_n) \bm K_{2g}\label{eq_update2},
\end{align}
where \(\bm K_{1g},\bm K_{2g}\) are maximum stiffness, the damping is set to be critical, i.e., the relationship between the \(j\)th diagonal entry of \(\bm D_i\) and \(\bm K_i\) is \(d_{ij}=2\sqrt{k_{ij}}\) for \(i=\{1,2\}\).

The desired positions \(\bm x_{1d},\bm x_{2d}\) are designed as follows. 
For {\em Waiting Mode}, $\bm x_{1d}, \bm x_{2d}$ is fixed to the pose at the moment entering this mode. Note that now the stiffness is high(all factors equal to 0), so the robot keeps still.
For {\em Scanning Mode}, $\bm x_{1d}, \bm x_{2d}$ are given by a scanning trajectory generator in our previous work \cite{yan2023multi}. 
For {\em Recovery Mode}, $\bm x_{1d},\dot{\bm x}_{1d},\ddot{\bm x}_{1d}$ are given by a mini-jerk trajectory generator, use controller (\ref{eq_tau1_tracking})(\ref{eq_tau2_regu}), $\bm x_{2d}$ is updated to actual robot value every 100 control cycles.
For {\em Human-Guiding Mode}, $\bm x_{1d}, \bm x_{2d}$ are set as current pose, i.e., $\bm x_{1d}(t) = \bm x_1(t), \bm x_{2d}(t) = \bm x_2(t)$). Note that in this mode, \(a_h\approx1\), the robot is soft and just follows the human.
For {\em Avoiding Mode}, \(\bm x_{1d}\) is the same with {\em Scanning Mode}, \(\bm x_{2d}{\mathrel{+}=}a_b \Delta\),
\(\Delta\in \Re\) is a fixed step length in the secondary task, if the doctor is on right \(\Delta>0\), left \(\Delta<0\). When the doctor is no longer close to the robot body, \(\bm x_{2d}\) will incrementally move back to its value when entering this mode.
For {\em Contacting Mode}, \(\bm x_{1d},\bm x_{2d}\) are the same with {\em Scanning Mode}. Note that now \(a_n = 1\), so the null-space stiffness is low, the robot will comply with the contact softly, without affecting the end effector scanning task.

\begin{algorithm}[!t]
\caption{Unified Interaction Control Algorithm}\label{alg_1}
\begin{algorithmic}[1]
\State progress time \(t_p=0\)
\While {$t_p < T$}
    \If{$a_h \ge a_{ht}$}  \label{alg_guiding_if}
        \State {\em Human-Guiding Mode}; break;
    \ElsIf{no trajectory} \label{alg_waiting_if}
        \State {\em Waiting Mode}; break;
    \ElsIf{have a trajectory $\{\bm x_d(i)\}_{i=1}^N$}\label{alg_traj_if}
        \State calculate the desired pose $\bm x_d(t_p)$ by (\ref{eq_calc_desired});
        \If{$(\lVert \tilde{\bm x}  \rVert<\epsilon)\&\&(a_p>a_{pt})||(a_f>a_{ft})$}\label{alg_traj_error_if}
            \If{\(a_b>a_{bt}\) and \(a_n>a_{nt}\)}
                \State {\em Contacting Mode}; break;
            \ElsIf{\(a_b>a_{bt}\) and \(a_n<a_{nt}\)}
                \State {\em Avoiding Mode}; break;
            \Else             
                \State {\em Scanning Mode}; break;
                \State $t_p = t_p+dt$;\label{eq_tp}
            \EndIf
        \Else 
            \State {\em Recovery Mode}; break;
        \EndIf
    \EndIf
\EndWhile
\end{algorithmic}
\end{algorithm}

The above explains how we achieve the ``Robot Behavior'' column in Tab. \ref{table_mode}  within one framework. Another part is how to decide the ``Activated When'' column. We show our decision method in Alg. \ref{alg_1}, where the subscript ``t'' denotes a user-defined threshold(e.g., \(a_h \ge a_{ht}\) means the factor is larger than its threshold). Variable \(t_p\hspace{-0.05cm}\in\hspace{-0.05cm}[0,T]\) denotes the task progress, where $T$ is the user-defined task time. At progress time $t_p$, the desired pose \(\bm x_d(t_p)=\bm x_d(i)\). The right term denotes the \(i\)th trajectory point, where
\begin{equation}
    i=\biggl[\frac{t_p\times N}{T}\biggr].\label{eq_calc_desired}
\end{equation}
This human-intention detection step can also be achieved in many ways like learning-based methods. 

\subsection{Passivity analysis}
In \cite{michel2022safety}, the authors augmented the hierarchical impedance controller with the energy tank method and rigorously proved the passivity property with variable impedance parameters and null-space projection. In this paper, we designed a specific time-varying way of the impedance parameters, allowing passivity to be proven similarly.
This key property enables our control framework to smoothly unify different working modes while maintaining a theoretical safety guarantee. 
In this way, our work is distinct from other works.
 
\section{Experiment}\label{sec_3d_vision}

\begin{figure} [tb]
    \centering
    \includegraphics[width=0.75\linewidth]{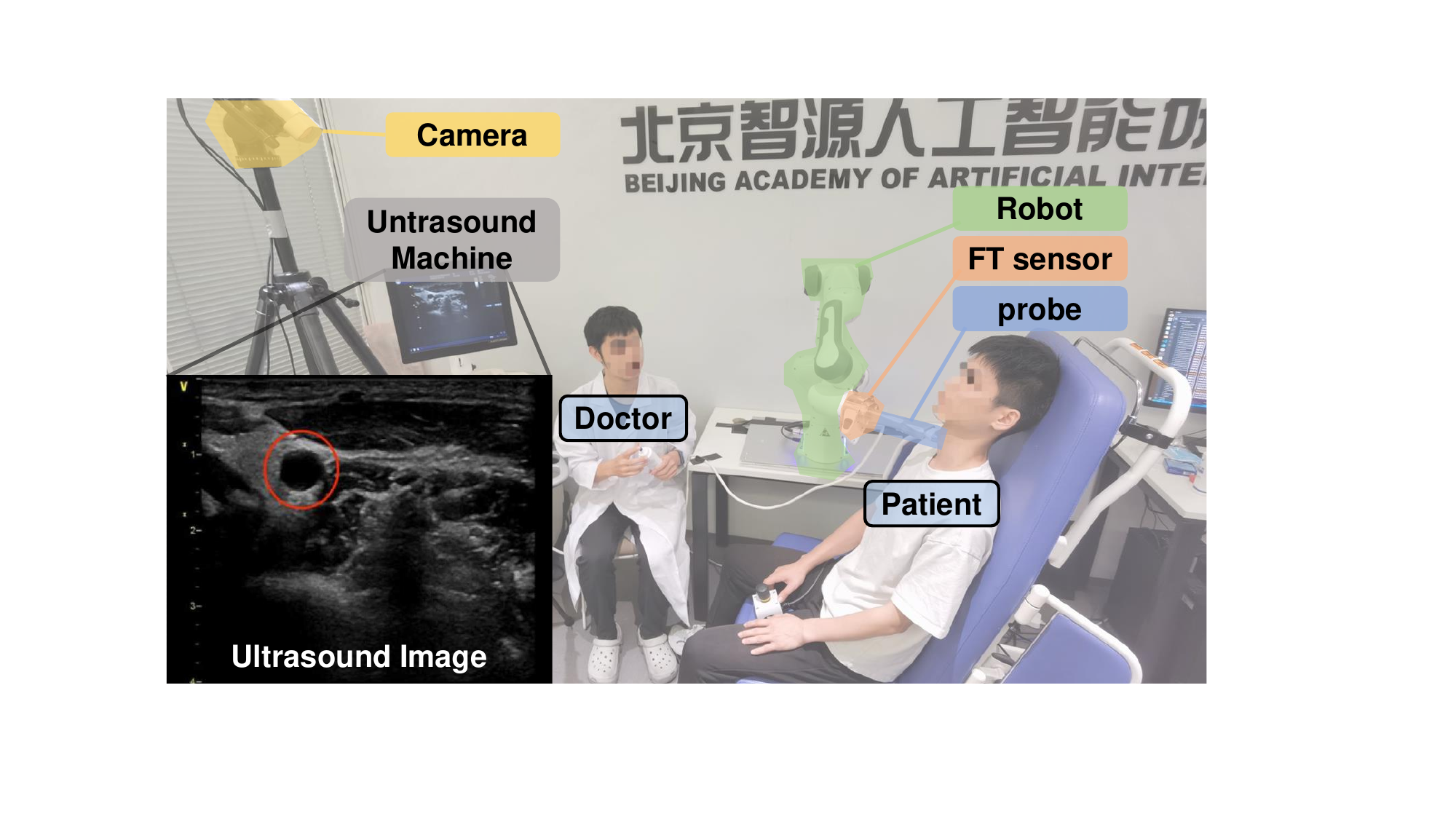}
    \vspace{-0.3cm}
    \caption{The experimental setup: (a) The hardware system consists of an ultrasound machine, an RGBD camera, a manipulator, an FT sensor, an ultrasound probe, and PCs; (b) An ultrasound image, where the red circle indicates the position of the cross-section carotid artery
    }
  \label{fig:exp_setup}
  \vspace{-6mm}
\end{figure}

Real-world experiments were carried out to validate the proposed control method. The experimental setup is shown in  Fig.~\ref{fig:exp_setup}.
The overall system consisted of six parts: (i) an ultrasound system (Vivid E7), including an imaging machine and a probe; (ii) a 7-DOF robot manipulator (Franka Panda); (iii) an ATI mini40 FT sensor mounted between the panda arm flange and the probe; (iv) an RGBD camera (Azure Kinect DK); (v) a hospital bed (which can be folded as a chair); and (vi) two PCs for processing the camera data and controlling the robot, respectively.

Using the RGBD camera, we employed the official body tracking algorithm to capture the patient's real-time neck frame and used point cloud data to generate a scanning trajectory through segmentation and reconstruction techniques. In the experiments, a human subject, simulating a patient, was seated on a hospital bed while a robot maneuvered a scanning probe along his neck to perform ultrasound imaging. A doctor was seated nearby, ready to intervene if needed.

Experiment 1 demonstrated the system's handling of intended interactions and its ability to transition between multiple interaction modes. Experiment 2 evaluated compliance with unintended interactions.

\subsection{Experiment 1 - Transition between Multiple Modes for Intended Interactions}
To simulate the complexity of the scanning task, we considered a restless patient who disrupts the process by dodging the probe, moving, sneezing, and deliberately pushing the probe. The doctor intervenes to manage unexpected situations. Stiffness varied, while damping was consistently set to be critical. Snapshots from different stages of the procedure are shown in Fig.~\ref{task1_snapshots}\footnote{Video at \url{https://yanseim.github.io/iros24ultrasound}\label{ft_video}}, and the weighting factors, stiffness, error and contact force are shown in Fig.~\ref{fig_task1_result}. The scanning task progressed as follows:
\begin{figure} [tb]
  \centering
    \includegraphics[width=0.9\linewidth]{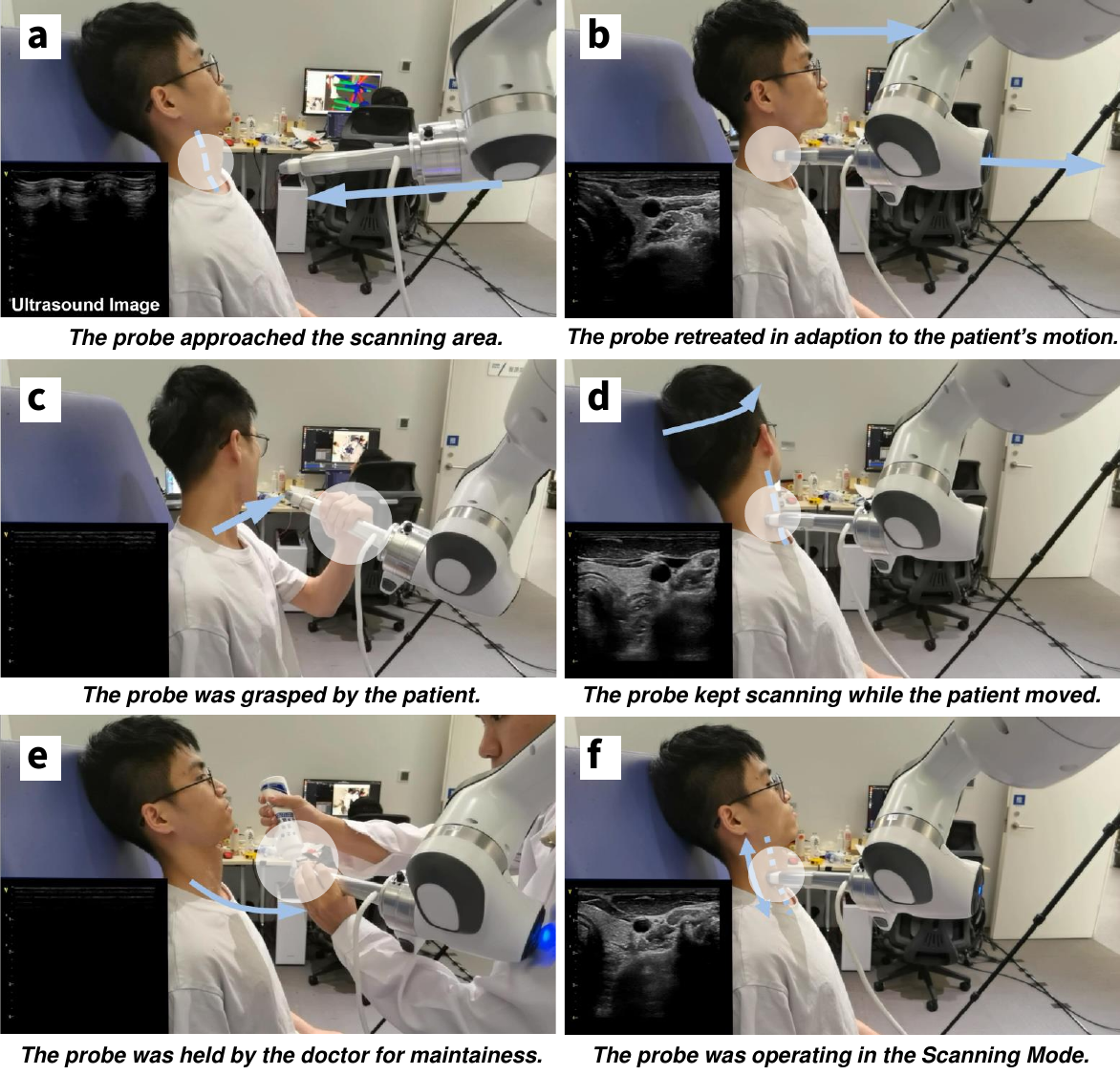}
  \vspace{-3mm}
  \caption{Snapshots of experiment 1: (a) \(t=11.5\mathrm{s}\). The probe was approaching the desired scanning area in the {\em Recovery Mode}; (b) \(t=37\mathrm{s}\). The probe retreated while the patient tilted forward; (c) \(t=55\mathrm{s}\). The probe was temporarily removed from the patient's neck; (d) \(t=68\mathrm{s}\). The probe followed the desired scanning trajectory while the patient turned to the side; (e) \(t=87\mathrm{s}\). The probe was held by the doctor for coupling gel to be applied; (f) \(t=130\mathrm{s}\). The probe was operating in the {\em Scanning Mode}.}
  \label{task1_snapshots}
  \vspace{-3mm}
\end{figure}
\begin{figure}[!t]
    \includegraphics[width=0.95\linewidth]{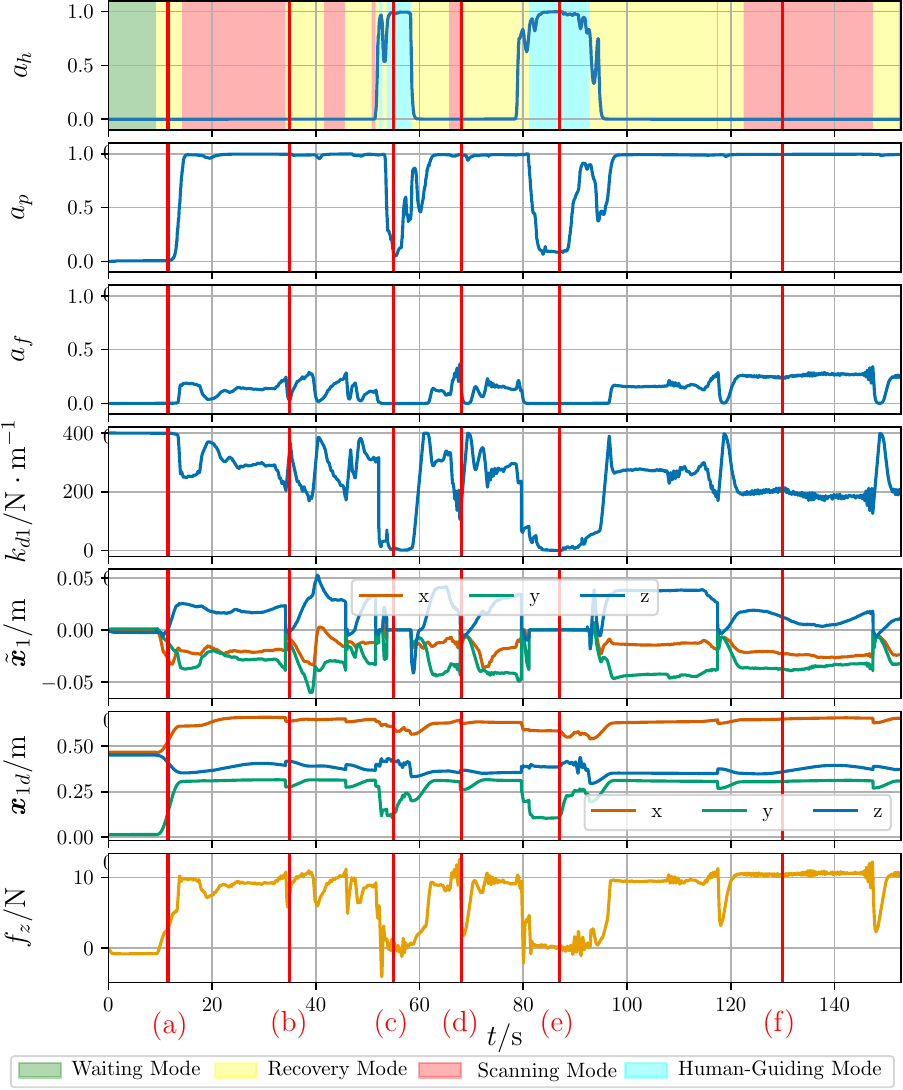}
    \vspace{-3mm}
 \caption{Experiment 1 results, from top to bottom: the first three subfigures correspond to the weighting factors of $a_h$, $a_p$, $a_f$ respectively; the fourth subfigure plots the translational stiffness of $k_{d1}$; the fifth subfigure plots the translational part of error $\tilde{\bm{x}}_{1}$; the sixth subfigure plots the desired translation of $\bm{x}_{1d}$; the seventh subfigure plots the estimated external force of $^{E}{f}_{z}$ expressed in end effector frame $\{E\}$. Background color in the first subfigure denotes different interaction modes, specifically, green - {\em Waiting Mode}, yellow - {\em Recovery Mode}, red - {\em Scanning Mode}, blue - {\em Human-Guiding Mode}. The red vertical lines denote instances in Fig. \ref{task1_snapshots}.
 }
 \label{fig_task1_result}
\vspace{-6mm}
\end{figure}
\begin{enumerate}
    \item The {\em Recovery Mode} activated between 10s and 15s, gradually increasing $a_p$ to 1, causing the scanning probe to approach the patient and establish physical contact (see Fig.~\ref{task1_snapshots}a).

    \item Between 35s and 45s, the patient pushed the probe forward twice, resulting in two peaks in $^{E}{f}_{z}$ and $a_{f}$. The increased tracking error $\tilde{\bm{x}_{1}}$ triggered {\em Recovery Mode}, followed by {\em Scanning Mode} (see Fig.~\ref{task1_snapshots}b). During this period, $a_{f}$ was crucial in attenuating $k_{d1}$ (\(\bm K_{1(0:3,0:3)}=k_{d1}\bm I\)), thus limiting the contact force $^{E}{f}_{z}$ and ensuring patient safety and comfort.

    \item During \qtyrange{51}{59}{\s}, the probe was manually moved away from the patient's neck (see Fig.~\ref{task1_snapshots}c), causing $k_{d1}$ to drop to \(0\) along with $a_h$, signaling the transition to {\em Human-Guiding Mode} where the robot became passive and responsive to manual guidance. A similar event occurred between 79s and 95s when the doctor applied coupling gel (see Fig.~\ref{task1_snapshots}e). Once the probe was released, $k_{d1}$ returned to its original value. These procedures showed a satisfactory reaction to intended interactions.

    \item At 66s, the patient turned his head to dodge the probe (see Fig.~\ref{task1_snapshots}d). The robot adapted and re-established contact at around 68s, reactivating {\em Scanning Mode}.

    \item The robot operated in {\em Scanning Mode} between 14s and 34s and again from 122s to 146s, maintaining stable contact with $a_p \approx 1$. The contact force was adequate for high-quality imaging and safety ($^{E}{f}_{z} \approx 10$ N, $a_{f} \leq 0.3$), with small tracking error ($\lVert\tilde{\bm{x}}_1\rVert \leq 0.04$ m). 
\end{enumerate}
Overall, the transitions were smooth and safety was ensured despite disturbances like patient movement or head turns. During the normal working period, the ultrasound images were clear, showing the vascular walls well. (see ultrasound images in Fig. \ref{task1_snapshots}).

\subsection{Experiment 2 - Compliance to Unintended Interactions}
\begin{figure} [!t]
  \centering
    \includegraphics[width=0.9\linewidth]{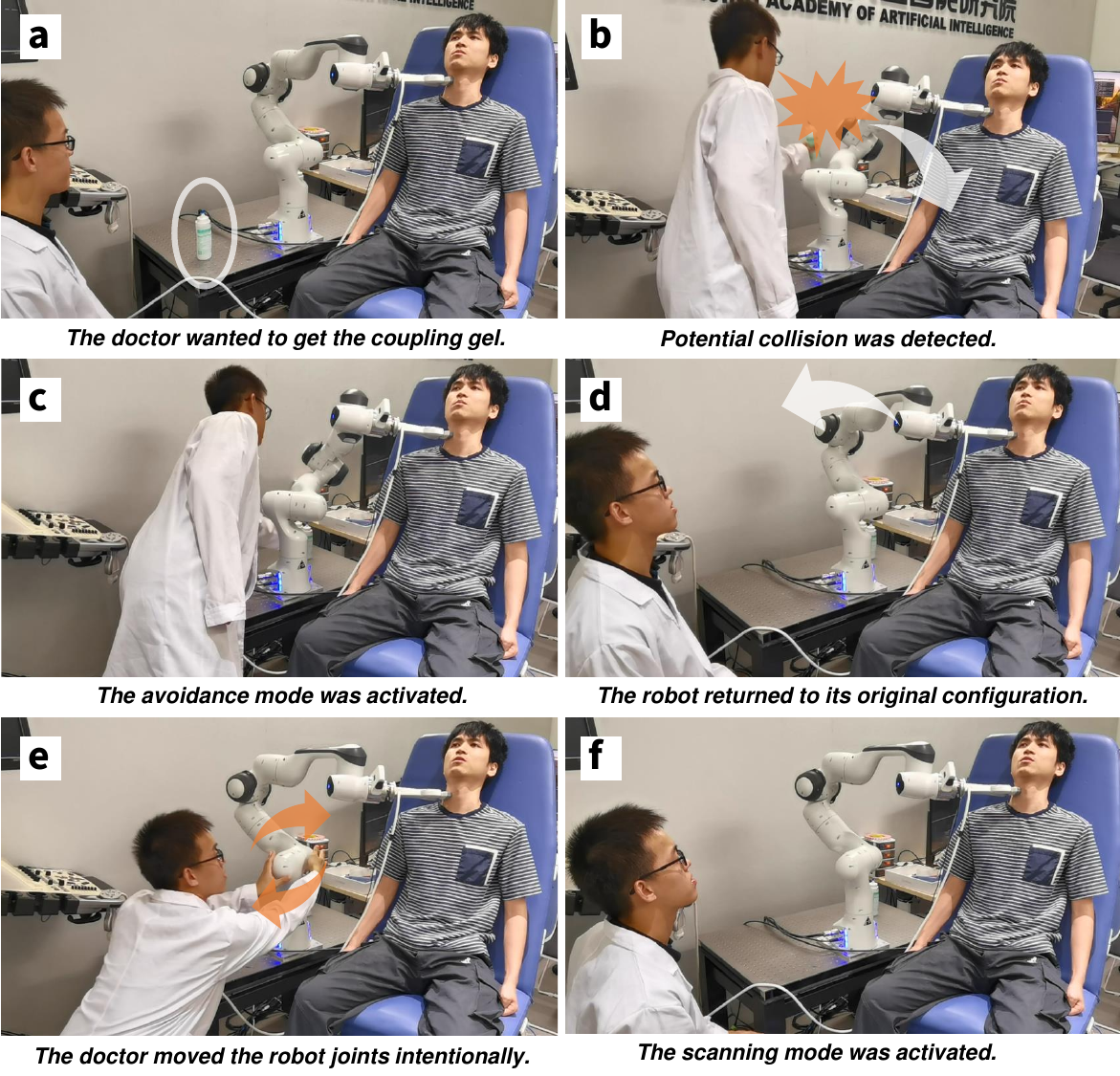}
  \vspace{-3mm}
  \caption{Snapshots of experiment 2: (a) \(t=2.5\mathrm{s}\). The robot was performing the scanning task. (b) \(t=5.2\mathrm{s}\). The doctor wanted to put the coupling gel on the back of the robot. The robot detected the potential collision and entered {\em Avoiding Mode}. (c) \(t=7.0\mathrm{s}\). The {\em Avoiding Mode} enabled the doctor to the place conveniently. (d) \(t=10.0\mathrm{s}\). The robot returned to its original configuration for scanning. (e) \(t=19.0\mathrm{s}\). The doctor contacted the robot in the null space, the {\em Contacting Mode} was activated, and the error and force of the end effector main task were unaffected. (f) \(t=25.0\mathrm{s}\). The {\em Scanning Mode} was activated.}\label{exp2_snapshots}
  \label{task2_snapshots}
  \vspace{-4mm}
\end{figure}

\begin{figure}[!t]
    \includegraphics[width=0.95\linewidth]{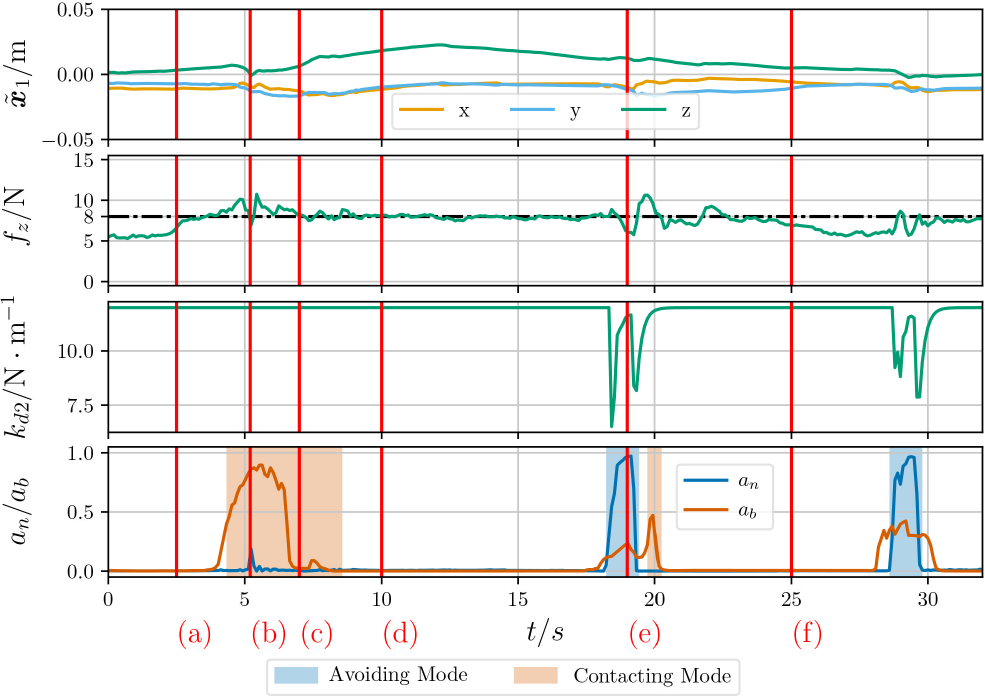}
    \vspace{-3mm}
 \caption{Experiment 2 results, from top to bottom: the position error of the main task; the contact force in the z direction of the end effector frame \(^{E}{f}_{z}\); the stiffness of the secondary task $k_{d2}$; weighting factors $a_n$, $a_b$. Background color in the last subfigure denotes different interaction modes, specifically, red - {\em Avoiding Mode}, blue - {\em Contacting Mode}. The red vertical lines denote instances in Fig. \ref{exp2_snapshots}.}\label{exp2-results}
 \label{fig_task2_result}
\vspace{-6mm}
\end{figure}

The snapshots of experiment 2 are shown in Fig. \ref{exp2_snapshots}. During the \(32\mathrm{s}\) scanning period, the doctor placed the coupling gel at around \(t
=5.2\mathrm{s}\), unintentionally activating {\em Avoiding Mode} (see the period in red in Fig. \ref{exp2-results}). As expected, there was no significant disturbance to the main task during {\em Avoiding Mode}, as seen in the \(\tilde{\bm x}_1\) and \(f_z\) figures. The null-space configuration then recovered as designed. 
Then, the doctor twisted the robot's body (Fig. \ref{task2_snapshots}e) to simulate an unintended collision at around \(t=18.5\mathrm{s}\). 
The stiffness of the secondary task dropped as in equation (\ref{eq_update2}) to be compliant with the collision and hence lower the collision force as much as possible.  
According to the patient, the force variation on the main task (See \(f_z\) in Fig. \ref{exp2-results}) was acceptable (See video). 
Note that the main task error and force were not affected by the null-space modulation no matter in {\em Avoiding Mode} or {\em Contacting Mode}, showing the effectiveness of the strict hierarchy and most importantly, guaranteeing the safety and comfort of the patient and the doctor. 

We conducted a user study on 9 male volunteers\textsuperscript{\ref{ft_video}}. The results showed satisfactory robustness of our developed system on different patients.  


\section{Conclusion}
The proposed unified interaction control framework allows the human to safely intervene at any time during the scanning to guide the robot's movement intentionally. It also allows the robot to avoid or partially accept unexpected collisions with humans by exhibiting compliance in null space without affecting the ongoing scanning task, ensuring patient safety. The system can distinguish between intended and unintended interactions, adapting the main task to the former and the null-space task to the latter. 
Experimental results from a carotid examination demonstrate that the system's human-intention-aware compliance ensures safety, comfort, convenience, and autonomous scanning, even with significant patient movement, doctor involvement, and unforeseen changes.
Our future work will be devoted to the development of more comprehensive human intention estimation approaches based on more types of human feedback.

\bibliographystyle{ieeetr}
\bibliography{main}

\end{document}